# THELXINOË: RECOGNIZING HUMAN EMOTIONS USING PUPILLOMETRY AND MACHINE LEARNING

Darlene Barker, PhD[1] and Haim Levkowitz, PhD[2]

[1]Department of Mathematics and Computer Science, Rivier University, Nashua, NH
[2]Miner School of Computer & Information Sciences,
Kennedy College of Sciences, University of Massachusetts, Lowell, MA

## ABSTRACT

*In this study, we present a method for emotion recognition in Virtual Reality (VR) using pupillometry. We analyze pupil diameter responses to both visual and auditory stimuli via a VR headset and focus on extracting key features in the time-domain, frequency-domain, and time-frequency domain from VR-generated data. Our approach utilizes feature selection to identify the most impactful features using Maximum Relevance Minimum Redundancy (mRMR). By applying a Gradient Boosting model, an ensemble learning technique using stacked decision trees, we achieve an accuracy of 98.8% with feature engineering, compared to 84.9% without it. This research contributes significantly to the Thelxinoë framework, aiming to enhance VR experiences by integrating multiple sensor data for realistic and emotionally resonant touch interactions. Our findings open new avenues for developing more immersive and interactive VR environments, paving the way for future advancements in virtual touch technology.*

## 1. INTRODUCTION

In a poetic sense, the eyes have long been regarded as the "window into the soul" offering a glimpse into the depths of human emotions and experiences [1]. In the realm of modern technology, this poetic vision transforms into a scientific reality, particularly in VR. The "pupils" serve as gateways not just "to the brain" but to the autonomic nervous system which subtly dilates and contracts in response to our emotions [1]. This study ventures beyond the realm of the soul, focusing on how the emotional state of a person influences the diameter of their pupils, and how this physiological response can be harnessed within the immersive world of VR [2].

In VR, where the digital and physical worlds converge, recognizing emotions can significantly enhance the overall experience, bringing it closer to the full spectrum of human interaction. This is where pupillometry, the measurement of pupil diameter, becomes a pivotal tool. Pupillary responses, which occur spontaneously and are challenging to control voluntarily, provide a continuous measure of emotional states, independent of a person's conscious awareness or control [1, 2, 3]. Leveraging this involuntary response, our research delves into the realm of VR, aiming to mirror real-life experiences by interpreting these subtle yet telling signs of our internal emotional landscape.

The emotions that we explored are the ones that "are related to the most basic needs of the body" like happiness, sadness, anger, and fear; of which the other emotions are composite [4]. Thelxinoe was an idea that started with the desire to accomplish the creation of the ability to touch another person across great distance whether via long-distance interactions or within a virtual space, to be able to feel the touch on the other side and feel feedback to that touch.

The idea is that you can manipulate objects in VR but not feel what you touch, or feel being touched. Our research is aimed at creating a framework to support this happening with the capturing of emotions of the parties involved using sensors, including cameras, to be able to capture the emotions from the brain, face, speech, and body movement [5].

The objective of our research was to see if the use of the pupil diameter was enough on its own to recognize human emotions using our emotion model of the four base emotions—happy, sad, anger and fear. The novelty of this approach is that we focussed solely on the pupil diameter and creating a working machine learning (ML) model with an accuracy of 98.8% using distinct emotions, and that these measures are "unconscious and difficult to control voluntarily" [6].

In this study, we explored emotion recognition with the use of pupil diameter. The initial data was collected from watching video content that matched the emotion model was and our approach using machine learning we classified the emotions with great accuracy. In our case, the stimuli were from video clips and with a non-invasive manner we used pupillometry to recognize the emotions [2]. We can control our emotions in the way we communicate verbally but we are not able to hide what we truly are feeling in our eyes. This is where pupillometry comes in, it provides "involuntary response" that we cannot control [7].

Anger and fear are ranging from 2.5 to 5mm while happiness is ranging from 2 to 5mm, and all emotions are all ranging between 2 and 5mm. For this reason, we explored additional features during the extraction process, and selection process, in preparation for classification.

"Pupillometry is the study of such changes in pupil diameter as a function of different neurological activities in a human brain" [7]. We studied pupillometry in recognizing emotions because of its dynamic nature compared to other physiological methods [8].

Another reason to study pupillometry is that our "learning, attention, perception, and memory is affected by our emotions" and by studying those emotions and their connections, we can move closer to recreating touch in VR [6]. This research is focused on the four basic emotions-fear, sadness, anger, and happiness [4].

## 1.1. Problem

We can see, hear, and manipulate objects within a VR experience. What we cannot do is feel when we run into a game object, or another player. Yes, there are currently ways to produce feedback with vibrations in the controllers. We require more than vibration; we need touch.

Because we respond to what we see more so than what we hear, Pupillometry is a necessary part of any emotion recognition frameworks such as Thelxinoë [3, 9].

Recognizing emotions in VR can bring us closer to a full body experience, and studying the senses that are already represented, we get closer to adding touch as part of the experience.

"Collecting emotions data from multiple sensors for Thelxinoë requires some aggregation of data to recognize the mental state of the user to process touch and generate the associated sensation quickly" [9]. At the same time, each sensor or sensor area needs to be studied separately and optimized independently of all the other sensors. In this case, we are studying the eyes as a sensor area independent to other sensor areas such as the brain, face, speech, and body movement.

This research is geared towards creating the framework necessary to be able to connect haptic devices to a human being and create an experience where touch occurs, and feeling is felt. We are aiming at putting the entire human body into the equation to be able to touch and be touched in VR.

## 1.2. Related Work

This method of recognizing human emotions from our eyes, from the subtle yet informative changes in pupil diameter as a window into the human emotional state is crucial to bring another sense into VR. A variety of machine learning models have been employed in recent studies to decode these physiological signals, each contributing uniquely to our understanding of the complex interplay between emotions and physiological responses.

Table 1 encapsulates a selection of seminal works in this domain, illustrating the diversity in methodologies and the varying degrees of accuracy achieved in classifying emotions. These studies use an array of ML model in classifying emotions.

| Study | Machine Learning Model Used | Accuracy |
|---|---|---|
| Four-class emotion classification in virtual reality using pupillometry [10] | Support Vector Machine (SVM) | 57.05% |
| | K-Nearest Neighbor (KNN) | 70.83% |
| Understanding the role of emotion in decision making process: using machine learning to analyze physiological responses to visual, auditory, and combined stimulation [11] | K-nearest Neighbors (KNN) | 52% |
| | Decision trees (DT) | 44% |
| | Logistic Regression (LR) | 51% |
| | Support Vector Machine (SVM) | |
| | Linear Discriminant analysis (LDA) | |
| | Random Forest (RF) | |
| | Adaboost (ADB) | |
| Machine Learning to Differentiate Between Positive and Negative Emotions Using Pupil Diameter [6] | K-nearest Neighbors (KNN) | 96.5% |
| Eye-Tracking Analysis for Emotion Recognition [13] | Support Vector Machine (SVM) | 80% |

Table 1: Some of the related works using Pupillometry to recognize human emotions.

One of the studies in Table 1 introduced a "technique to detect and differentiate between positive and negative emotions" with a 96.5% accuracy [6]. We show in figure 1 the taxonomy of emotions showing them falling into either neutral, or non-neutral which covers all other emotions that are either positive or negative. This was the closest to what we studied but still different in that we used a distinct emotion model.

A study of cognitive Pupillometry methods that provides "a comprehensive," firsthand "guide to methods in cognitive Pupillometry, with a focus on trial-based experiments in which the measure of interest is the task evoked pupil response to stimulus" as well as expressed the need for a system for processing Pupillometry data [15].

Another study argues "that Pupillometry is a powerful and non-invasive tool to investigate emotional processing in clinical populations" [8]. While this is not physically invasive, it is emotionally invasive. It is assumed that the users of the framework that this study is a part of consent to the exchange of information about each other.

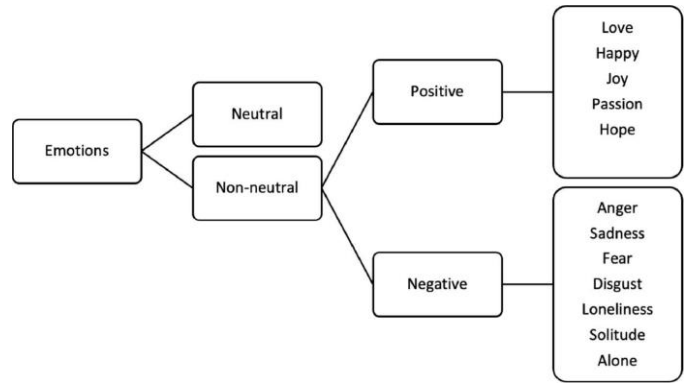

Figure 1: The taxonomy of emotions from [14].

### 1.3. Motivation

This paper is part of our research for the creation of a framework called Thelxinoë [89]. This technology will bring the whole body into VR to achieve a close to whole-body experience. The framework will allow those in a VR world to touch another person, feel this sensation, and have the other person feel it as well. In this work, we focused on the pupils of the eyes and what emotions we can recognize using the pupil diameters during the experience of the basic emotions.

Replacing in-person interactions is not the goal here but to simulate them. With the added information that one would not have in an in-person interaction, this information gotten from sensors we can get a better measure of what the person is feeling which can narrow the distance or allow for greater immersion in a virtual space. Knowing this information would analogously characterize touch as in-person touch, without seeing either as being a real-life interaction vs a manufactured experience [12]

## 2. METHODS AND MATERIALS

This research includes one of the key areas of recognizing human emotions, the eyes, as a source of the data of study. Here we look at how our body responds to external stimuli to trigger the changes in pupil size. The autonomic nervous system controls the changes in your pupil whether it be light in your environment or an emotional state, dilating when happy or excited and constricting when sad. It is hard to hide our emotions when it expressed in the changes in the pupil.

"Pupillometry is a powerful and" physically non-invasive "tool to investigate emotional processing" making this "a promising tool in" research [1, 8]. This part of the Thelxinoë framework covers emotion recognition in our "pupil diameter" to better understand the emotional state of the user [1, 3, 8, 13]. While pupil constriction and dilation are assumed to be a reaction to negative and positive stimuli, it has been found that "pupil dilation is far more dynamic" [8].

During this research, we used an emotion model that contains the four basic emotions: anger, sadness, happiness, and fear. The data that is collected during the collection phase will go through the pre-processing phase, followed by the feature extraction, then the feature selection phase, classification, and finally the statistical analysis phase.

One thing that is key while studying pupillometry is that pupil dilation is a measure of emotion, cognitive load, and memory recognition but also the objective amount of light that the eye is exposed to [16]. The light was controlled by maintaining a constant light source in the room where the datasets were collected—only light changes in the video content were used as stimuli. Breaks were taken between each emotional state that was triggered or type of stimuli, to reduce any potential cognitive overload.

Memory recognition was not addressed since the assumption in this work is that a happy video would evoke happy emotions and somehow remind the user of happy memories, the same with the other basic emotions.

## 2.1. Data Collection

While "eye tracking is a powerful tool that can be applied to a wide variety of research questions" in our approach, we only used the diameter of the pupil resulting from dilation and contraction in response to stimuli, for the recognition of the emotional state of the user [17]. "The most primary metric used in Pupillometry is the pupil diameter" validated our choice [18].

We collected data using the HP Reverb G2 Omnicept Edition headset, as seen in Fig. 3, within a VR environment with video content serving as stimuli that evoke emotions, happiness, sad, anger, and fear.

As seen in Fig. 2 our approach involves data collection, pre-processing, feature extraction, feature selection, classification, and statistical analysis of the results, to recognize the emotions from the data we collected of pupil diameter data.

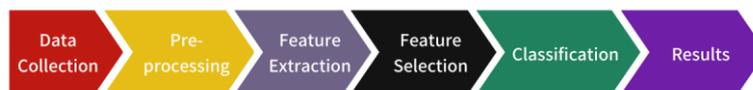

Figure 2: The flow of the emotion recognition pipeline used.

We used the pupil diameter of both eyes and collected for 10-minute periods for each emotion, happiness, sadness, anger, and fear. This data went through pre-processing to remove noise, feature extraction to gain more information about the dataset, feature selection to choose the optimal features for the dataset, classifying the samples into one of the emotions, and finally the statistical analysis to show how well the model performed.

## 2.2. Sensors

The device used in the research was the HP Reverb G2 Omnicept Edition VR Headset shown in fig. 3. It contains an Eye Tracking sensor. We used Unity3D 2021.3.19f1 Personal for the environment that we created for data collection. The other software that we used in this research were MATLAB R2022b, Python 3.11.2, and HD Omnicept SDK.

For data collection, we created a space where the user could not move since all they needed to do was start an executable of the scene created in Unity 3D and put on the headset. The world included a large screen where the video clips played. We attempted to limit movement and anything that would limit results, such as changes in lighting in the room where the data was collected. We also captured results from both eyes. Data was captured from watching video clips from shows that were expected to create the desired emotions in the user.

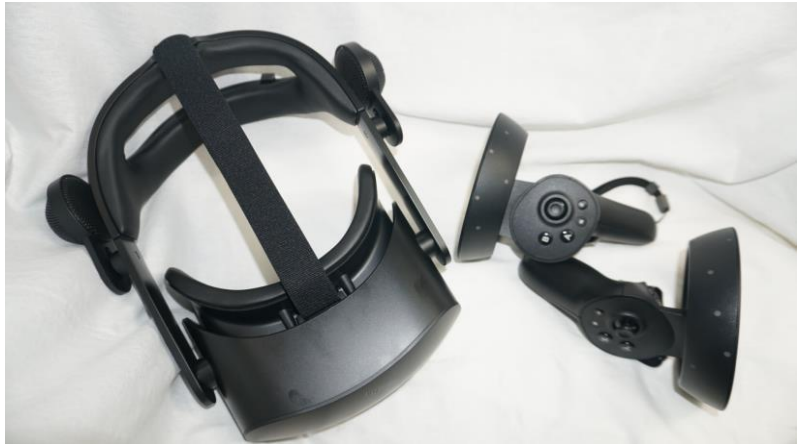

Figure 3: HP Reverb G2 Omnicept Edition

We focused solely on pupil diameter and ignore gaze and cognitive overload because the tests intended to see if it was possible to classify emotions using the pupil diameter [6]. We ignored where the person was looking on the screen as well as saccades, which is the rapid eye movement between fixation points on the screen. Each session was stored in a file with the emotion trigger in the name.

## 2.3. Data Preprocessing

We created the datasets we used from the viewing in Unity3D of video content which were the source triggers for each emotion. The design matrix contained the timestamp of the data collection, left-eye pupil diameter, and right-eye pupil diameter as shown in figure 4. As part of the preprocessing step, the dataset of all four emotions was compiled using the name of the file to create the label of each record held in the file, also as shown in Figure 4.

Figure 4: Sample of eye tracking data showing pupil sizes for happy.

The pupil diameter range for the sensor is 2-8mm, any value small is reported in the data as -1 which is representative of blinks. We removed the noise in the form of blinks. We also removed samples where one eye was closed, which usually either led up to or followed a blink event [19]. The other pre-processing that we carried us was replacing the date and time with a timestamp.

In [3], Nakakoga et al. used principal component analysis (PCA) to remove "artifacts" in the pupillary data. In our data gathered from the Omnicept SDK, blinks are recorded as -1 and we used that behavior to remove artifacts from the data. All data that contained a blink representative of the -1 on one eye or both eyes, were removed from the raw dataset before feature engineering.

They start with a clean dataset of the dilation measurements of the left and right eyes while viewing emotional stimuli within a virtual environment. The individual emotion data samples were labelled and combined to form the dataset of all four of the emotions. This raw dataset consists of a timestamp, two features (left eye, right eye), and a label, like that shown in Fig. 4 with the happy label.

## 2.4. Feature Extraction

The term "VR-generated" in the context of our abstract refers to data that is produced or collected within a Virtual Reality (VR) environment. It implies that the data, particularly the pupillometry data is gathered while participants are engaged in a VR setting, experiencing visual and auditory stimuli through a VR headset.

In VR, stimuli can be highly controlled and varied in ways that might not be possible in a real-world setting. This allows for the collection of detailed and specific data under a range of controlled conditions, which can be particularly valuable for research in fields like emotion recognition, behavioral studies, and human-computer interaction.

In this step, we used MATLAB to extract additional features from the two features: leftEye, and rightEye. This process included statistical features (mean, standard deviation, covariance, and kurtosis) as well as the power spectral density (PSD) [19]. Kurtosis was used to compute the sample kurtosis for the input blocks providing an estimate of the shape of the data distribution's tail. This is useful for finding outliers, or extreme values in the dataset, and these extreme pupil dilations or constrictions are crucial in the different emotional states.

The extraction process was done on 120 block samples. The dataset included 193 additional columns 18 columns were removed because they were zeros, dropping down to 175 columns.

## 2.5. Feature Selection

The feature selection process used a combination of mRMR and GridSearchCV to go from 175 to 50 optimal features that we used for our classification model.

The initial feature set began with 175 features that were extracted from the data. This large set of features included both highly informative and less relevant or redundant features. We applied the mRMR technique to find a subset of features that would have maximum relevance as well as those features that had minimum redundancy. The goal here was to minimize the potential of overfitting.

We reduce the number of extracted features from 175 to 50 features and with further reduction use GridSearchCV to find the top optimal features [20, 21, 22,23, 24]. The decision to reduce the number of features to fifty was based on an analysis of feature importance, as illustrated in Figure 5. Feature importance scores help identify which features contribute most significantly to the model's predictive ability. Here we could observe that the importance of features drops to zero around the fifties indicating that beyond this point, additional features do not meaningfully contribute to model performance.

Further experimentation, such as testing with the top thirty features, was conducted to validate the choice of fifty features. It was found that reducing to thirty features led to a drop in accuracy, indicating that while some reduction is beneficial, overly aggressive feature pruning can remove valuable information about the data.

Figure 5: The feature importance graph of fifty-eight features while experimenting with GridSeasrchCV.

The overall goal of feature selection is to find a balance between model complexity and performance. Too many features can lead to overfitting and increased computational cost, while too few features might fail to capture important nuances in the data.

## 2.6. Classification

Before the classifier used the dataset, the samples were shuffled. For each dataset we used they were "divided into a training set (70%) and a test set (30%)" [11]. Like other papers we read, we used SVM but chose a GB based off the stacking of Decision Trees with greater accuracy. We performed tuning we used "grid search" to achieve the "best hyperparameters for" the GB model, as shown in fig. 6. We used a DecisionTreeClassifier as the base estimator of the GB model along with subsampling set to 0.8 for better generalization, meaning the model being able to adapt to previously unseen data [24].

```python
from sklearn.metrics import mean_squared_error
gbrt = GradientBoostingClassifier(max_depth = 5,
                                  learning_rate=0.05,
                                  n_estimators=20,
                                  max_features=7,
                                  min_samples_split=200,
                                  min_samples_leaf=30,
                                  subsample=0.8,
                                  random_state=10)
gbrt.fit(X_train, y_train)

errors = [mean_squared_error(y_test,y_pred)
          for y_pred in gbrt.staged_predict(X_test)]
bst_n_estimators = np.argmin(errors)

gbrt_best = GradientBoostingClassifier(loss='deviance',
                                       criterion='friedman_mse',
                                       max_depth = 5,
                                       learning_rate=0.051,
                                       max_features=7,
                                       min_samples_split=200,
                                       min_samples_leaf=30,
                                       subsample=0.8,
                                       random_state=10,
                                       n_estimators=bst_n_estimators)
gbrt_best.fit(X_train,y_train)
```

Figure 6: GB model including the Hyperparameters.

Another model that we used was the Long Short-Term Memory (LSTM) with an accuracy of 61-70% without further finetuning. This was not pursued further because the Pupillometry data will be combined with the EEG data in future works and processed together. This decision was based upon the accuracy achieved by another work that used combined signals and achieved similar accuracy using a similar model as well as the need to address a multimodal approach to process the data from multiple sensors in real-time in future research of Thelxinoë [9, 25].

## 2.7. Results

Our GB model generated 85% accuracy without feature engineering, for both training and test datasets. For datasets that underwent feature engineering, the accuracy was 95% for both training and test datasets. In addition to accuracy, we achieved the following: specificity (99%), sensitivity (98%), precision (98%), f-score (98%), and Matthew Correlation Coefficient (97%), shown in Figure 7 below:

```
Acc ---Overall accuracy for each class = 0.9798001275781416
Spec --Specificity or true negative rate =  0.9926596651234642
Rec ---Sensitivity, hit rate, recall, or true positive rate (macro) = 0.9805812912400057
Pres --Precision or positive predictive value =  0.9844670223042051
F =  0.9821526841500154
Matthew Correlation Coefficient, MCC = 0.9727902047235911
```

Figure 7: Additional Metrics for our GB Model.

In figure 8, shows the confusion matrix of our GB model which achieved a 98.8% accuracy.

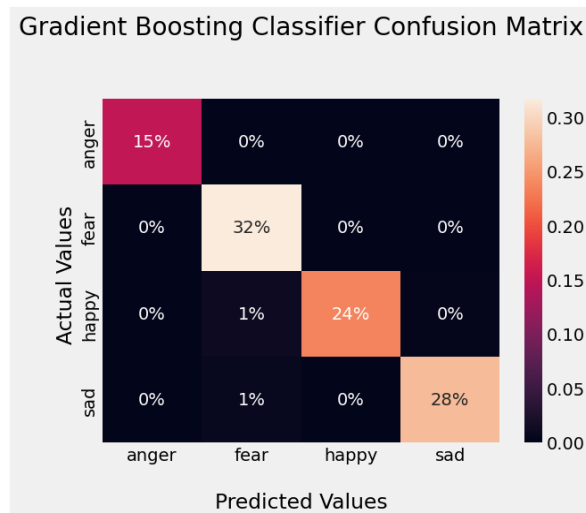

Figure 8: Gradient Boosting Machine Classifier Confusion Matrix

## 3. DISCUSSION

In our study, the initial feature set f or each sample comprised timestamps and pupil diameters for both the left eye (LE) and right eye (RE). Interestingly, when identifying the most crucial features, we observed a predominance of data from the LE. Out of the top features, most were derived from the LE, with only seven features representing the RE, as detailed in Fig. 9 for the top thirty features for clarity.

This finding echoes the work of Rafique et al. who focused on identifying a range of emotions based on pupil diameter, using data from the left eye [7]. Our approach expanded on this by using both eyes, yet our results similarly highlighted a greater significance in the LE data. Notably, our feature extraction process revealed that the majority of the top fifty-one features were LE-centric, with figure 6 illustrating the distribution.

We exclusively relied on pupil diameter measurements, excluding confidence levels from our data collection. This approach yielded successful results, with accuracies of 84.9% without feature engineering and 98.8% with.

The environmental setup for our data collection was intentionally controlled and minimalistic, with fixed lighting and no additional interactions in the VR space, apart from a static background. This design choice was to ensure data consistency and reduce variables that could potentially influence pupil responses.

Our Gradient Boosting (GD) model demonstrated impressive accuracies, consistently scoring in the mid-to-high nineties. However, the Long Short-Term Memory (LSTM) offline model showed lower performance, with accuracy in the mid-to-high sixties. In future research phases, we plan to explore the integration of LSTM in processing aggregated data. Nevertheless, the current GD model stands as a robust tool for recognizing emotions through pupillometry, offering promising results and potential for further refinement.

The GD model was able to show accuracies in the mid-to-high nineties but with offline model LSTM currently coming in in the mid-to-high sixties. We were able to achieve high eighties to high nineties with the same model on the EEG data. We decided that the use of LSTM for the aggregated data to be addressed when the aggregation phase of the research happens but for now, we have a model for recognizing pupillometry data using GD with satisfactory results.

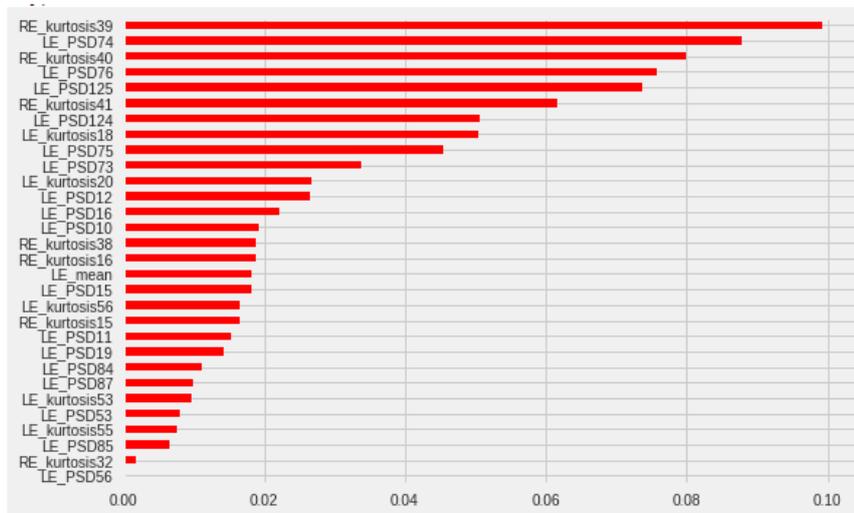

Figure 9: The 30 top features while experimenting with GridSearchCV.

## 3.1. Future Work

One obvious step would be to aggregate the EEG and pupillometry data in recognizing emotions [17]. In doing this, the datasets are collected on different sample frequencies, 256Hz, and 120Hz, for EEG, and Pupillometry, respectively. The sampling change will need to be addressed. For the offline model, the LSTM model will need to be further finetuned [26].

Expand the emotion model to include more than the basic emotions after addressing the multimodal emotion recognition.

Addressing the study of dominance in eyes and if this would affect the emotion recognition process. Most of the optimal features were from the left eye extracted features although the right eye gets the number one spot while the feature spread seems to change. This brings us to question whether studying both eyes is of significance to the recognition of emotions, in one of the studies we looked at, the pupil diameter was the same for both eyes but with our research that was not the case, hence the use of both [7].

Another would be potentially "examining the full waveform of pupil activity, from resting baseline, to constriction, to" re-dilation to see if it affects accuracy we have already achieved [27].

Yet another approach would use eye movement to further understand what the user is thinking. The user looking away along, combined with an emotional state could explain their behavior within the virtual world. One of the studies found that "eye movement such as fixation and saccaded depend largely on the dynamics of the movie scene" [9]. So, for our research after the physical environment and other integrations this could be useful information, which will be part of body language, and a more controllable emotional response.

Finally, it is important to develop safeguards to protect participants using Thelxinoë. This would mean conducting research into what measures to put in place and how to put them. In cases where the user is not experiencing the expected emotions one will need to detect if there is an "affective deficit" [8]. One study suggested that "psychopathy is associated with a problem in processing negative emotions such as sadness and fear" [8].

## 4. CONCLUSIONS

In this research study, we focused on recognizing emotions using pupillometry data within a VR environment. The primary goal was to determine whether using just the pupil diameters as features in response to the visual and auditory stimuli, would we be able to accurately recognize human emotions.

The research included feature extraction techniques, including time-domain, frequency-domain, and time-frequency domain composite features, extracted from datasets captured in VR. These features were processed to reduce the dimensionality and selecting the top features through feature engineering. An ensemble learning model that stacked multiple decision trees and used gradient boosting, was used to achieve emotion recognition.

The study achieved an accuracy of 98.8% after feature engineering, and model tuning. Even without feature engineering, the model achieved 84.9% accuracy. This outcome demonstrated the effectiveness of using pupillometry data for emotion recognition.

The contribution of this work is to add recognition of emotions from pupil diameters to the Thelxinoë framework, aimed at enabling touch sensations in VR interactions by collecting data from various sensors to generate meaningful emotions. This work represents a foundational step towards enhancing immersion and emotional interaction within virtual spaces.

# REFERENCES

[1]   N. Ferencova, Z. Visnovcova, L. B. Olexova, and I. Tonhajzerova, "Eye pupil– a window into central autonomic regulation via emotional/cognitive processing," Physiological Research, vol. 70, no. Suppl 4, p. S669, 2021.

[2]   S. G. G. Laeng, Bruno; Sirois, "Pupillometry," Perspectives on Psychological Science,     vol.    7, no. 1, pp. 18–27,     2012. [Online]. Available: https://browzine.com/articles/15295768

[3]   S. Nakakoga, H. Higashi, J. Muramatsu, S. Nakauchi, and T. Minami, "Asymmetrical characteristics of emotional responses to pictures and sounds: Evidence from pupillometry," PloS one, vol. 15, no. 4, p. e0230775, 2020.

[4]   S. Gu, F. Wang, N. P. Patel, J. A. Bourgeois, and J. H. Huang, "A model for basic emotions using observations of behavior in Drosophila," Frontiers in psychology, p. 781, 2019.

[5]   D. Barker and H. Levkowitz, "Thelxinoë for Emotion Recognition and Touch Processing," 2023 IEEE International Conference on Advanced Systems and Emergent Technologies (IC_ASET), Hammamet, Tunisia, 2023, pp. 01-06, doi: 10.1109/IC_ASET58101.2023.10150503.

[6]   A. Babiker, I. Faye, K. Prehn, and A. Malik. "Machine Learning to Differentiate Between Positive and Negative Emotions Using Pupil Diameter," Frontiers in Psychology, vol. 6, 2015.

[7]   S. Rafique, N. Kanwal, I. Karamat, M. N. Asghar, and M. Fleury, "Towards Estimation of Emotions from Eye Pupillometry with Low-Cost Devices," IEEE Access, vol. 9, pp. 5354–5370, 2021.

[8]   D. T. Burley, N. S. Gray, and R. J. Snowden, "Emotional modulation of the pupil response in psychopathy." Personality Disorders: Theory, Research, and Treatment, vol. 10, no. 4, p. 365, 2019.

[9]   D. F. Barker. "Creation of Thelxinoë: Emotions and Touch in Virtual Reality." Doctoral dissertation, University of Massachusetts Lowell, 2023.

[10]  Zheng, Lim Jia, James Mountstephens, and Jason Teo. "Four-class emotion classification in virtual reality using pupillometry." Journal of Big Data 7 (2020): 1-9.

[11]  E. M. Polo, A. Farabbi, M. Mollura, L. Mainardi, and R. Barbieri. "Understanding the role of emotion in decision making process: using machine learning to analyze physiological responses to visual, auditory, and combined stimulation." Frontiers in human neuroscience vol. 17 1286621. 8 Jan. 2024, doi:10.3389/fnhum.2023.1286621.

[12]  A. Gallace, and C. Spence, In Touch with the Future: The sense of touch from cognitive neuroscience to virtual reality (Oxford, 2014; Oxford Academic, 16 Apr. 2014)

[13]  P. Tarnowski, M. Kolodziej, A. Majkowski, and R. J. Rak, "Eye-tracking analysis for emotion recognition," Computational intelligence and neuroscience, vol. 2020, 2020.

[14]  N. Ahmed, Z. A. Aghbari, and S. Girija, "A systematic survey on multimodal emotion recognition using learning algorithms," Intelligent Systems with Applications, vol. 17, p. 200171, 2023. [Online]. Available: https://www.sciencedirect.com/science/article/pii/S2667305322001089

[15]  A. Mathˆot, Sebastiaan; Vilotijeviʹc, "Methods in cognitive pupillometry: Design, pre-processing, and statistical analysis," Behavior Research Methods, 2022. [Online]. Available: https://browzine.com/articles/533468444

[16]  S. Rafique, N. Kanwal, I. Karamat, M. N. Asghar, and M. Fleury, "Towards estimation of emotions from eye pupillometry with low-cost devices," IEEE Access, vol. 9, pp. 5354–5370, 2020.

[17]  B. T. Carter and S. G. Luke, "Best practices in eye tracking research," Inter- national Journal of Psychophysiology, vol. 155, pp. 49–62, 2020.

[18]  D. Shadiev, Rustam; Li, "A review study on eye-tracking technology usage in immersive virtual reality learning environments," Computers & Education, vol. 196, p. 104681, 2023. [Online]. Available: https://browzine.com/articles/538583070

[19]  H. Sharma, L. Drukker, A. T. Papageorghiou, and J. A. Noble, "Machine learning-based analysis of operator pupillary response to assess cognitive workload in clinical ultrasound imaging," Computers in Biology and Medicine, vol. 135, p. 104589, 2021. [Online]. Available: https://www.sciencedirect.com/science/article/pii/S0010482521003838

[20]  J. Zhang, Z. Yin, P. Chen, and S. Nichele, "Emotion recognition using multi- modal data and machine learning techniques: A tutorial and review"," Information Fusion, vol. 59, p. 103–126.

[21]  X. Li, Y. Zhang, P. Tiwari, D. Song, B. Hu, M. Yang, Z. Zhao, N. Kumar, and P. Marttinen, "EEG based emotion recognition: A tutorial and review," ACM Computing Surveys, vol. 55, no. 4, pp. 1–57, 2022.

[22] M. R. Islam, M. A. Moni, M. M. Islam, M. Rashed-Al-Mahfuz, M. S. Islam, M. K. Hasan, M. S. Hossain, M. Ahmad, S. Uddin, A. Azad, et al., "Emotion recognition from EEG signal focusing on deep learning and shallow learning techniques," IEEE Access, vol. 9, pp. 94 601–94 624, 2021.

[23] E. H. Houssein, A. Hammad, and A. A. Ali, "Human emotion recognition from EEG-based brain–computer interface using machine learning: a comprehensive review," Neural Computing and Applications, pp. 1–31, 2022.

[24] A. Natekin and A. Knoll, "Gradient boosting machines, a tutorial," Frontiers in neurorobotics, vol. 7, p. 21, 2013.

[25] S. Vijayakumar, R. Flynn, P. Corcoran, and N. Murray, "BiLSTM-based Quality of Experience Prediction using Physiological Signals," in 2022 14th International Conference on Quality of Multimedia Experience (QoMEX), 2022, pp. 1–4.

[26] S. Rafique, N. Kanwal, M. S. Ansari, M. Asghar, and Z. Akhtar, "Deep Learning based Emotion Classification with Temporal Pupillometry Sequences," in 2021 International Conference on Electrical, Computer and Energy Technologies (ICECET), 2021, pp. 1–6.

[27] A. I. Mckinnon, N. S. Gray, and R. J. Snowden, "Enhanced emotional response to both negative and positive images in post-traumatic stress disorder: evidence from pupillometry," Biological Psychology, vol. 154, p. 107922, 2020.

## AUTHORS

**Haim Levkowitz** is the Chair Emeritus of the Computer Science Department at the University of Massachusetts Lowell, in Lowell, MA, USA, where he has been a faculty member since 1989. He was a twice-recipient of a US Fulbright Scholar Award to Brazil (August – December 2012 and August 2004 – January 2005). He was a Visiting Professor at ICMC — Instituto de Ciencias Matematicas e de Computacao (The Institute of Mathematics and Computer Sciences)—at the University of Sao Paul, Sao Carlos – SP, Brazil (August 2004 - August 2005; August 2012 to August 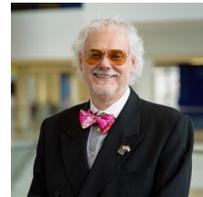
2013). He co-founded and was Co-Director of the Institute for Visualization and Perception Research (through 2012). He is a world-renowned authority on visualization, perception, color, and their application in data mining and information retrieval. He is the author of "Color Theory and Modeling for Computer Graphics, Visualization, and Multimedia Applications" (Springer 1997) and co-editor of "Perceptual Issues in Visualization" (Springer 1995), as well as many papers in these subjects. He is also co-author/co-editor of "Writing Scientific Papers in English Successfully: Your Complete Roadmap," (E. Schuster, H. Levkowitz, and O.N. Oliveira Jr., eds., Paperback: ISBN: 978-8588533974; Kindle: ISBN: 8588533979, available now on Amazon.com: http://www.amazon.com/Writing-Scientific-Papers-English-Successfully/dp/8588533979). He has more than 48 years-experience teaching and lecturing, and has taught many tutorials and short courses, in addition to regular academic courses. In addition to his academic career, Professor Levkowitz has had an active entrepreneurial career as Founder or Co-Founder, Chief Technology Officer, Scientific and Strategic Advisor, Director, and venture investor at several high-tech startups.

**Darlene Barker** is an Assistant Professor of Computer Science at Rivier University in the Mathematics and Computer Science department. She has a PhD in Computer Science from the University of Massachusetts Lowell, in Lowell, MA, USA. She completed her master's degree in computer science from University of Massachusetts. She also completed her Bachelor of Science majoring in Computer Science from Rivier University in Nashua, NH. Her professional career experience spanning several years and disciplines from systems to software working for companies—Compaq 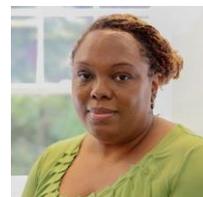
Computer Corporation, Teradyne, Inc., and Oracle Corporation. She is the Founder of an indie game studio, DB Attic Studios, LLC, in Bedford, NH where she makes video games with her children from middle school through high school graduates.